\documentclass[sigconf,nonacm]{acmart}

\AtBeginDocument{%
  }

\acmConference[KDD '26]{Proceedings of the 32nd ACM SIGKDD Conference on Knowledge Discovery and Data Mining}{August 9--13, 2026}{Jeju, South Korea}
\acmBooktitle{Proceedings of the 32nd ACM SIGKDD Conference on Knowledge Discovery and Data Mining (KDD '26)}
\usepackage{booktabs}
\usepackage{array}

\begin{document}

\title[AI-Native 6G: A BlueSky Vision]{Towards Resilient and Autonomous Networks:\\ A BlueSky Vision on AI-Native 6G}
\titlenote{Accepted at the 32nd ACM SIGKDD Conference on Knowledge Discovery and Data Mining (KDD '26).}
\author{Liang Wu, Kelly Wan, Mayank Darbari, and Liangjie Hong}
\email{{liang.wu, liangjie.hong}@nokia.com}
\affiliation{%
  \institution{Nokia}
  \city{Sunnyvale}
  \state{California}
  \country{USA}
}

\renewcommand{\shortauthors}{Wu et al.}

\begin{abstract}
The proliferation of emerging applications, such as autonomous driving and immersive experiences, demands cellular networks that are not only faster, but fundamentally more resilient and autonomous. This paper presents a BlueSky vision on how Artificial Intelligence will be natively integrated into 6G, shifting the paradigm from \underline{Network for AI} to \underline{AI for Network}. We envision that, unlike 5G's reliance on scattered, ad-hoc models each trained for a single task, native AI in the 6G era will be anchored by a foundation model and and orchestrated via collaborative multi-agent systems, framing network management as a unified, multi-modal, multi-task optimization problem. Built on this vision, we outline two transformative directions. The first focuses on developing a 6G foundation model as a unified backbone, with task-specific knowledge distilled into compact models suited for diverse edge deployments. The second advances multi-agent systems designed to autonomously diagnose, maintain, and recover networks with minimal human intervention. These directions chart a roadmap for 6G to evolve into an intelligent, self-sustaining communication infrastructure.
\end{abstract}

% \keywords{Do, Not, Use, This, Code, Put, the, Correct, Terms, for, Your, Paper}
\maketitle

\section{Introduction}
The integration of machine learning into cellular networks has demonstrated compelling gains throughout the 5G and 5G-Advanced era~\cite{shahid2025large}. Task-specific models have been shown to improve a wide range of network functions, including traffic prediction, beamforming optimization, and handover anticipation. Yet as we look toward 6G, the demands placed on wireless infrastructure extend far beyond transmitting bits faster from one point to another. Emerging applications such as autonomous driving, unmanned aerial vehicles, and intelligent wearables introduce qualitatively new requirements. For example, networks must sense device locations, infer states of cellular network users, and respond to failure modes that have no precedent in prior generations~\cite{tataria20216g, chataut20246g}. Addressing each requirement by training a dedicated, ad-hoc model, as has been the prevailing practice in 5G-Advanced~\cite{chen20235g}, leads to a fragile and costly ecosystem. The proliferation of narrow models inflates development overhead, complicates maintenance, and introduces subtle incompatibilities when the output of one model serves as input to another. More fundamentally, a collection of siloed, single-objective models cannot coherently balance competing network goals—throughput, latency, energy efficiency, and user experience—when these objectives must be jointly optimized in real time.

To address these challenges, we propose a vision centered on two complementary pillars. The first is a 6G foundation model: a unified, multi-modal model capable of jointly processing heterogeneous inputs—natural language, raw wireless signals, positioning data, and sensory streams from connected devices. Rather than maintaining a fragmented library of task-specific models, this foundation model serves as a shared knowledge backbone from which task-specific behaviors can be derived through fine-tuning, and efficient, deployment-ready variants can be obtained through knowledge distillation tailored to the resource constraints of diverse edge devices. This design consolidates semantic understanding across modules, reduces development and maintenance costs, and ensures coherent inter-module communication. The second pillar is a multi-agent system that translates the foundation model's intelligence into coordinated network action. Individual agents operate in parallel across network domains—access, core, and transport—handling maintenance, configuration, and recovery tasks autonomously. The two pillars constitute our vision for AI-native 6G: a system where a unified foundation model resolves the challenge of knowledge integration, and a multi-agent architecture resolves the challenge of action coordination, jointly enabling networks that are at once more resilient and more autonomous. Table~\ref{tab:5g_vs_6g} summarizes the key differences between 5G and our proposed 6G AI-native vision across three dimensions. Unlike 5G, which treats AI as an add-on for isolated tasks, 6G natively integrates a unified foundation model and autonomous multi-agent systems, pivoting the paradigm from 'Network Supporting AI' to a fundamentally 'AI-Native Network'.

\begin{table}[t]
\centering
\caption{Comparison of 5G and the Proposed 6G AI-Native Vision}
\label{tab:5g_vs_6g}
\renewcommand{\arraystretch}{1.3}
\begin{tabular}{@{}p{1.5cm}p{2.9cm}p{2.9cm}@{}}
\toprule
\textbf{Aspect} & \textbf{5G} & \textbf{6G} \\
\midrule
Connectivity
  & Devices connect for communication only; no network-level spatial awareness
  & Pervasive sensors enable real-time 3D localization of UAVs, autonomous vehicles, and mobile users \\
\midrule
AI Model
  & Siloed, task-specific models (e.g., separate models for traffic prediction, beamforming, handover)
  & Unified communication foundation model supporting multi-modal, multi-task optimization \\
\midrule
Operations
  & Human-as-doer: engineers manually diagnose faults and configure network parameters
  & Human-as-supervisor: autonomous multi-agent systems handle diagnosis, maintenance, and recovery across RAN and core \\
\bottomrule
\end{tabular}
\end{table}

\section{Vision: A 6G Foundation Model}
A truly AI-native 6G network demands a new class of model—one that is not assembled from a library of task-specific modules, but is instead built from the ground up to reflect the diversity, scale, and criticality of telecommunications workloads. We envision a 6G foundation model organized around three defining properties. First, it must be multi-modal, capable of natively ingesting and jointly reasoning over the heterogeneous data streams that define modern wireless networks: raw signal measurements, time-series telemetry, and spatiotemporal traces from connected devices. Second, it must be multi-task, able to perform not only the forecasting tasks familiar from prior work, but also anomaly detection, root cause analysis, and human-readable explanatory reasoning—all within a unified model. Third, it must support efficient adaptation, serving as a general-purpose backbone that can be rapidly specialized, compressed, or distilled to meet the stringent latency and resource constraints of diverse deployment contexts, from core network controllers to the most constrained edge devices. The three properties define not merely an incremental improvement over existing models, but a qualitative rearchitecting of how intelligence is embedded in the network.

\subsection{Multi-Modal Telecom Understanding}

The dominant paradigm in telecom-oriented foundation models today is to adapt large language models to the telecommunications domain through fine-tuning on domain-specific corpora—technical standards, configuration logs, and operational manuals~\cite{qu2025mobile, chen2025llm, maatouk2024large, zhou2024large}. While this approach yields models with useful linguistic fluency about network concepts, it remains fundamentally a language modeling exercise. Such models are ill-suited to the core demands of AI-native 6G, which center not on text comprehension, but on the real-time interpretation of raw network signals, sensor streams, and spatiotemporal data that are native to the wireless medium.

Time-series foundation models (TSFMs) represent a more promising point of departure, yet existing benchmarks on telecommunications datasets reveal a consistent performance gap: models trained predominantly on meteorological, financial, or IoT sensor data fail to transfer effectively to telecom workloads~\cite{feng2025telecomts}. Two structural factors explain this gap. The first is a data mismatch: public TSFM training corpora contain negligible quantities of telecommunications-origin signals, and the statistical properties of network traffic—bursty, non-stationary, and driven by the superposition of many simultaneous user behaviors—differ markedly from the low-dimensional, smooth time series these models were designed for~\cite{ansari2024chronos, ansari2025chronos, liu2025moirai, cohen2024toto}. The second is an architectural mismatch: standard TSFM tokenization strategies, such as patch-based segmentation~\cite{ansari2024chronos, ansari2025chronos, liu2025moirai, cohen2024toto}, were designed for signals that are well-approximated by a small number of latent factors (temperature, humidity, seasonal effects). Telecommunications data is inherently high-rank: a traffic matrix aggregated across a cell sector reflects the simultaneous behavior of hundreds to thousands of users, each subject to independent behavioral dynamics. Patch-based tokenizers collapse this structure, discarding the cross-variate correlations that carry diagnostic information~\cite{feng2025telecomts}. A telecommunications foundation model therefore requires new architectural primitives: tokenizers designed for high-rank, high-frequency multivariate signals; attention mechanisms that explicitly model covariate relationships across antenna ports or frequency bands; and decoding strategies capable of producing both autoregressive outputs and quantized discrete predictions, as emerging approaches such as Chronos-2 have begun to explore.

Beyond time-series telemetry, the 6G ecosystem introduces an additional layer of modality complexity. As 6G networks integrate dense sensor infrastructure to support autonomous vehicles~\cite{he20206g}, unmanned aerial systems~\cite{shrestha20216g}, and extended reality applications~\cite{yu2023toward, ahmad2023leveraging}, the foundation model must reason over three-dimensional spatiotemporal data—positional traces, velocity fields, and mobility patterns—whose geometry bears no resemblance to one-dimensional time series. These trajectories unfold in physical space, carry explicit coordinate structure, and demand architectural treatment suited to their spatial semantics, whether through geometric encoders, graph-based representations of device neighborhoods, or coordinate-aware attention. A 6G foundation model that cannot natively process this spatiotemporal modality will be systematically blind to the most distinctive AI requirements of the new generation, which emphasizes on connecting edge devices with remote sensors focusing on locations and trajectories.

\subsection{Multi-Task Telecom Reasoning}

Existing foundation models for time series are largely designed around a single canonical objective: predicting the next time step or the next patch of time steps~\cite{kottapalli2025foundation}. This framing captures prediction accuracy as the primary success criterion, and it is well-suited to domains where forecasting is the end goal. Telecommunications, however, is a domain where forecasting is rarely sufficient. The network operations center does not merely want to know what traffic will look like in the next ten minutes; it needs to know when something has gone wrong, why it went wrong, and what should be done about it.

This broader task space demands a model that is simultaneously anomaly-aware during training and capable of structured causal reasoning at inference time. The challenge is particularly acute because telecommunications time series are inherently high-dimensional: a single network incident may manifest as coordinated deviations across dozens of correlated metrics spanning multiple layers of the protocol stack~\cite{736100ffac29485b9dbfbe01dc7c625d}. In 5G networks, such fault diagnosis already taxes human operators; in 6G, where the data volume and source diversity grow substantially, it becomes practically intractable without automated support. We envision a 6G foundation model trained with explicit supervision over both normal operational patterns and failure signatures, enabling it to not only flag deviations but to trace their propagation across dependent subsystems and generate human-readable explanations of likely root causes. This shift from accurate prediction to interpretable causal narration is essential to realizing the 6G design philosophy of moving from human as doer to human as supervisor~\cite{pang2026toward}.

The multi-task imperative extends beyond time-series telemetry to the spatiotemporal modalities described above. When the foundation model monitors the trajectories of unmanned aerial vehicles~\cite{yang2024deep} or the mobility patterns of autonomous vehicles~\cite{baccari2024anomaly}, it must simultaneously maintain predictive models of nominal behavior and anomaly detectors sensitive to deviations that may signal mechanical failure, adversarial interference, or environmental disruption. These tasks are not independent: accurate trajectory forecasting improves anomaly sensitivity, and anomaly detection provides feedback that refines the generative model. A unified foundation model architecture that couples these objectives—rather than maintaining separate models for prediction and detection—naturally captures this interdependence and supports the kind of joint optimization that high-stakes 6G applications require.

\subsection{Efficient Adaptation and Deployment}

The consolidation of diverse capabilities into a single foundation model offers substantial efficiency benefits in development and maintenance, but it introduces a complementary challenge: the model must be rapidly and reliably specialized to serve specific operational contexts, and it must be made lightweight enough to function on the heterogeneous hardware landscape that characterizes real 6G deployments~\cite{zeeshan2025knowledge}.

Specialization requirements arise across two dimensions. First, operational scenarios may demand heightened precision for particular task types: a core network controller managing high-value enterprise traffic may require anomaly detection sensitivity far exceeding baseline performance, while an edge node serving a dense urban venue may prioritize latency-sensitive load forecasting. Rather than retraining from scratch, we envision capability-specific fine-tuning pipelines that can selectively amplify a targeted capacity—automatically curating the training signal most informative for that capability, or applying structured fine-tuning that updates only the parameters most responsible for it. This second path, in turn, connects to the broader program of interpretability for neural network models: understanding which circuits within the foundation model implement which functions makes it possible to route fine-tuning updates to where they will have the greatest effect~\cite{leireforming}, and to prune functionality that is irrelevant or counterproductive in a given deployment context~\cite{zhang2025flexible}.

Second, inference efficiency is a first-class constraint in telecommunications. Many network functions operate under strict latency budgets that a large foundation model cannot meet without modification. For reasoning-intensive tasks such as root-cause analysis, this requires fine-tuning the model toward more concise reasoning chains—fewer tokens, lower computational cost, no sacrifice in answer quality~\cite{pedroso2025anomaly}. For continuous, high-throughput monitoring tasks at the edge, the relevant mechanism is knowledge distillation: extracting a compact student model from the foundation model backbone, targeted at a specific capability rather than attempting to replicate the full model's generality. Edge devices in a 6G network are not interchangeable; they differ widely in compute, memory, and the nature of the raw data streams they observe. The distillation pipeline must therefore be automated and capability-aware, producing a family of deployment-specific models each optimized for its local task, resource budget, and data regime. In the most aggressive case, rather than distilling a new model end-to-end, it may be possible to extract and prune specific functional subnetworks from the foundation model directly, yielding lightweight modules that retain a targeted capability while discarding the rest—a form of surgical model decomposition that becomes tractable only when the model's internal structure is sufficiently understood.

\section{Vision: Multi-Agent Systems for Autonomous 6G Networks}
Throughout the 5G and 5G-Advanced era, machine learning has been progressively integrated into network operations—yet the role of human operators has remained central~\cite{larsson20255g, nam2014advanced}. Routine tasks such as fault triage~\cite{hu20205g}, configuration updates~\cite{pozza2020reconfiguring}, and capacity planning~\cite{perez2015artificial} still require direct human execution, while only a narrow set of ML-assisted functions have advanced to a state where humans serve as guides rather than doers. The result is a network management paradigm that is partially automated but fundamentally human-dependent. The vision we advance here is more ambitious: by deploying coordinated multi-agent systems across both the radio access and core network domains, 6G can achieve a state in which humans serve primarily as supervisors. That is only setting goals and overriding decisions in exceptional cases, while the network autonomously handles the vast majority of operational, maintenance, and recovery tasks. This section articulates how that vision translates into concrete agent architectures on each side of the network, and identifies the shared knowledge infrastructure required to support both.

\subsection{RAN-Side Multi-Agent Systems}
A distinguishing hardware characteristic of 6G base stations is the transition from fixed-function, application-specific processing units toward general-purpose, scalable GPU-based compute platforms~\cite{basaran2025next, nokianews}. Unlike their predecessors, these platforms support elastic capacity expansion and are natively suited to the batch processing workloads that neural inference demands. Crucially, this architectural shift dissolves a longstanding bottleneck: rather than transmitting raw data from the device to a remote compute facility for processing, a significant portion of inference and decision-making can now be executed locally at the base station, eliminating the associated transmission latency and backhaul cost. This co-location of computation and radio access creates the foundation for responsive, low-latency agent operation, but it simultaneously introduces a new class of coordination challenge that did not exist when compute and radio were physically separated.

The most immediate coordination requirement is task routing. A compute offloading agent must continuously evaluate how inference workloads should be distributed across the three available tiers: user equipment, base station, and remote cloud or core, balancing latency constraints against energy budgets and real-time compute availability. This routing decision is neither static nor independently solvable. It depends on the current channel state, the nature of the task, the mobility profile of the device, and the load on upstream compute resources, all of which evolve on timescales of milliseconds to seconds.

Mobility management in 6G extends substantially beyond the handover prediction~\cite{lima2023deep} familiar from prior generations. As 6G networks integrate dense sensing infrastructure to support autonomous vehicles, unmanned aerial systems, and extended reality devices, a dedicated sensing and localization agent must process Integrated Sensing and Communication (ISAC)~\cite{liu2022survey} data streams to maintain real-time estimates of device position, velocity, and trajectory within the three-dimensional deployment space. These estimates feed directly into mobility decisions and anomaly detection, and they represent a qualitatively new input modality that prior handover models were not designed to consume.

Classical radio resource management~\cite{agarwal2022comprehensive} tasks are similarly transformed by agentic treatment. A beamforming agent, rather than optimizing beam weights from instantaneous channel measurements alone~\cite{ahmed2018survey}, can draw on a persistent context window comprising user mobility history, predicted trajectory, and long-term channel statistics to produce personalized, anticipatory beam configurations that improve link quality under mobility conditions where reactive approaches degrade. A spectrum and power management agent must simultaneously determine how spectrum resources~\cite{ravi2025spectrum} are allocated across concurrent tasks of differing criticality, prioritizing the low-latency control channel of an autonomous vehicle over a background file transfer. These decisions are interdependent. For example, power levels will affect interference, interference may affect spectrum efficiency, and spectrum allocation constrains what beam configurations can be feasible.

As the number of specialized agents grows, inter-agent coordination itself becomes a first-order problem. Individual agents optimize local objectives; without a mechanism for arbitration, their actions can conflict in ways that degrade global network performance. A spectrum agent reducing interference may inadvertently starve a beamforming agent of the resources it requires; a power-saving agent may conflict with a coverage assurance agent during a sudden demand surge. Resolving these conflicts requires a RAN orchestrator agent that maintains a global view of agent states and objectives, detects emerging conflicts, and adjudicates among competing actions using a principled priority ordering aligned with network-level goals. Such hierarchical multi-agent architectures for telecommunications represent an open research frontier: no deployed system currently implements this form of coordinated RAN autonomy, and realizing it will require advances in multi-agent reinforcement learning—both for training individual task agents and for designing orchestrators capable of stable coordination under dynamic, partially observable conditions. Federated learning will play an equally important role, enabling agents distributed across base stations to collaboratively improve shared models without centralizing sensitive user data.

\subsection{Core Network-Side Multi-Agent Systems}
The core network is the functional heart of a cellular system~\cite{parvez2018survey}, responsible for authenticating devices, managing sessions, enforcing policies, routing traffic to its destination, and exposing network capabilities to external services. In 5G, these functions are implemented as a set of loosely coupled, cloud-native network functions—the Access and Mobility Management Function (AMF), Session Management Function (SMF), Policy Control Function (PCF), Unified Data Management (UDM), and others—each handling a distinct slice of the control plane. While this service-based architecture introduced valuable modularity, its operation remains heavily human-supervised: slice provisioning, charging configuration, policy updates, and fault resolution all require substantial operator involvement. The 6G core network, as we envision it, replaces this human-in-the-loop model with a layer of autonomous agents that manage each functional domain independently and coordinate across domains through structured orchestration.

Session management~\cite{park2022session} is a natural first target for agent deployment. In a 6G network carrying simultaneously the traffic of consumer broadband, autonomous vehicle telemetry, industrial IoT control loops, and immersive media, sessions differ not only in their quality-of-service requirements but in their failure semantics, latency tolerances, and billing models. Dedicated session management agents—specialized respectively for data, voice, and control traffic—can maintain awareness of these distinctions and act on them in real time: renegotiating session parameters in response to changing conditions, preemptively migrating sessions ahead of predicted congestion, and escalating to the orchestration layer when cross-session tradeoffs must be resolved.

Charging and subscription management~\cite{chen20235g}, historically among the most labor-intensive and error-prone aspects of network operations, are particularly well-suited to agent-based automation. A charging agent continuously monitors usage patterns against active subscription terms, detects billing anomalies—whether arising from misconfiguration, equipment failure, or fraudulent behavior—and autonomously triggers corrective actions such as personalized alerts, adaptive rate limiting, or billing adjustments, without requiring a human operator to review each event. The shift from periodic, batch-oriented billing reconciliation to continuous, real-time agent monitoring represents a meaningful improvement in both operational efficiency and user experience.

Network slicing~\cite{zhang2019overview}, which refers to the virtualization of the physical network into isolated logical instances tailored to the requirements of different customers or applications, is a 5G capability that has not yet been operationalized at scale, largely because provisioning, monitoring, and lifecycle management of slices still demand significant manual effort. In 6G, a slice lifecycle agent can autonomously instantiate new slices in response to service requests, scale capacity in anticipation of demand surges, enforce SLA compliance through continuous KPI monitoring, and decommission slices when they are no longer needed. This closes the loop between business intent and network configuration in a way that manual processes cannot match at 6G timescales.

As on the RAN side, the plurality of core network agents necessitates a core orchestrator agent that coordinates across functional domains, resolves policy conflicts—for instance, between a slice SLA agent demanding more compute and an energy efficiency agent constraining it—and maintains end-to-end service coherence across the RAN-core boundary. Federated learning and distributed inference techniques will be equally relevant here, enabling models that span multiple operator domains or geographic regions to be trained collaboratively while preserving data sovereignty.

\subsection{Supporting Knowledge Infrastructure}
Realizing the multi-agent architectures described above requires two foundational infrastructure components that do not yet exist in deployable form. The first is a network digital twin~\cite{nguyen2021digital}, \textit{i.e.}, a continuously updated, generative simulation of the live network that serves as a training environment for reinforcement learning agents and a sandbox for evaluating proposed actions before they are applied to production systems. A digital twin, in the telecommunications context, is more than a static topology model: it is a dynamic, data-driven simulator that ingests real-time telemetry from across the network and maintains a high-fidelity replica of network state, including traffic distributions, interference patterns, device mobility, and failure propagation dynamics. Agents trained against a faithful digital twin can acquire robust policies without incurring the cost and risk of learning directly from live network interactions. Generating such a twin at the fidelity required for 6G, encompassing heterogeneous access technologies, NTN integration~\cite{saleh2025integrated}, and dense sensing modalities, is itself an open research challenge, not limited to wireless networks, one that will demand generative modeling techniques capable of capturing the non-stationary, multi-scale dynamics of real deployments.

The second infrastructure component is a telecommunications knowledge graph~\cite{krinkin2020models} that encodes the structured relationships among network entities, configurations, failure modes, and remediation procedures. Where the foundation model provides statistical pattern recognition over raw signals, the knowledge graph provides symbolic, interpretable structure that agents can query through retrieval-augmented generation to ground their decisions in domain knowledge. A knowledge graph that links, for instance, a particular alarm signature to its known root causes, the network functions involved, and the recommended resolution procedures enables agents to reason over failure scenarios that may be rare in the training data but well-documented in operational knowledge bases. The digital twin and the knowledge graph constitute the epistemic foundation on which autonomous 6G agents must be built—one providing experiential knowledge through simulation, the other providing knowledge through structured representation.

The proliferation of agentic AI across both network domains, however, introduces a new attack surface and a new class of privacy risks that the architectures described above do not yet fully address. We turn to these concerns in the following section.

\section{Security and Privacy of AI-Native 6G}
% The long-term trajectory of 6G security points toward a fundamental transformation: as quantum communication technologies mature, the physical layer of 6G networks will offer information-theoretic security guarantees that no classical cryptographic system can match. Quantum key distribution and post-quantum cryptographic primitives promise a communication substrate that is, in principle, naturally resistant to eavesdropping and forgery. Yet this vision remains a horizon. In the near and medium term, the transition to AI-native 6G introduces a substantially expanded attack surface—one shaped not by classical cryptographic vulnerabilities, but by the novel properties of foundation models, multi-agent coordination, and the pervasive sensing infrastructure that 6G demands. The security and privacy challenges of this transition period deserve careful attention precisely because they are distinct in kind from those of prior generations, and because the consequences of failure in a highly autonomous, self-managing network are qualitatively more severe.

% \subsection{Security Challenges of the 6G Foundation Model}
The shift from a fragmented ecosystem of task-specific models to a unified foundation model resolves many of the coordination and maintenance problems identified in prior sections—but it simultaneously concentrates risk in a way that has no precedent in 5G deployments. In 5G, a compromised or corrupted model affects a single task in a bounded operational context. In 6G, the foundation model is the shared cognitive substrate from which all agent behaviors are derived; a vulnerability in the foundation model is therefore a vulnerability in the entire network.

%Training data poisoning is a known threat in machine learning, but its consequences in the 6G context are qualitatively different from those in isolated 5G deployments. When a foundation model is trained on poisoned data—whether through deliberate injection by a malicious participant in a federated training pipeline, or through corruption of a shared telemetry repository—the resulting bias is not confined to a single downstream task. %Because the foundation model serves as a shared backbone for anomaly detection, root-cause analysis, beamforming optimization, and session management simultaneously, a targeted poisoning attack can be designed to degrade a specific capability while leaving others intact, evading detection by conventional monitoring. Worse, the distillation pipeline that propagates foundation model knowledge to edge devices amplifies the reach of such an attack: a single poisoned training run can silently corrupt every lightweight model deployed across the network's edge. Defending against this threat requires new techniques for cross-task poisoning detection—methods that can identify training samples whose influence is disproportionately concentrated in a targeted subset of the model's capabilities, rather than uniformly distributed across tasks.

A defining feature of the 6G foundation model, as envisioned in this paper, is its ability to receive and act on natural language instructions from network operators and, through intent-based interfaces, from enterprise customers. This capability is precisely what enables the shift from human-as-doer to human-as-supervisor—but it also introduces a class of vulnerabilities with no analog in prior network management systems. Prompt injection attacks, in which adversarial instructions are embedded within seemingly legitimate inputs, can redirect the foundation model's behavior in ways that are difficult to detect through conventional input validation. In the context of network management, a successfully injected prompt does not merely produce a harmful text output; it can translate directly into a network reconfiguration, a slice provisioning decision, or a traffic rerouting action that affects millions of users. The natural language interface that makes the foundation model accessible and powerful is the same interface that makes it exploitable, and securing it requires advances in input sanitization, behavioral constraint enforcement, and anomaly detection over the model's action space rather than its output text alone.

In 5G, location information is a derived quantity, inferred from signal strength measurements with limited spatial resolution. In 6G, precise three-dimensional position is a first-class data type, continuously produced by the sensing and localization agents and ingested directly into the foundation model's spatiotemporal reasoning pipeline. The foundation model that can predict an autonomous vehicle's trajectory with high accuracy can, by the same mechanism, maintain persistent behavioral profiles of individual users—profiles that reveal home and work locations, daily routines, social associations, and movement anomalies. The precision that makes 6G sensing valuable for safety-critical applications is the same precision that makes it dangerous as a surveillance instrument.

Conventional access control models, including role-based and attribute-based access control frameworks, were designed for environments in which access decisions are made at authentication time and remain stable for the duration of a session. Autonomous agents present a fundamentally different access control problem: an agent's required permissions change dynamically as its task context evolves, and the appropriate scope of its access at any moment depends on what it is currently trying to accomplish and on the current state of the network. The challenges define a security and privacy research agenda that is as ambitious as the AI-native network architecture it must protect, which is the foundation for all data-driven applications in 6G including AI and ML. These challenges define a security and privacy research agenda that is as ambitious as the AI-native network architecture it must protect, which is the foundation for all data-driven applications in 6G including AI and ML. A slice lifecycle agent that legitimately requires write access to provisioning interfaces during a scale-out operation should not retain that access during normal monitoring phases. %A charging agent that needs to read session records for anomaly detection should not be able to modify them. Enforcing these context-sensitive boundaries requires a new class of dynamic, intent-aware access control mechanisms—ones that grant permissions based on the agent's declared task and current operational phase, enforce automatic expiration when the task context changes, and log access decisions in a tamper-evident audit trail that supports post-hoc accountability. Extending attribute-based access control to this agentic setting is a meaningful research challenge, requiring formal models of agent intent, task lifecycle, and permission scope that do not yet exist in deployable form.

% % The unification that makes 6G powerful—a shared foundation model, a coordinated agent fabric, a pervasive sensing infrastructure—is the same unification that amplifies the consequences of any single point of failure. Meeting this challenge will require not merely the application of existing security techniques to new contexts, but the development of fundamentally new approaches to trustworthy autonomy in critical infrastructure.

\section{Conclusion}
This paper has presented a BlueSky vision for AI-native 6G, arguing that the limitations of 5G's fragmented, task-specific ML paradigm cannot be resolved by incremental refinement, but only by a fundamental rearchitecting of how intelligence is embedded in the network. We have proposed two complementary pillars toward this end. The first is a 6G foundation model—multi-modal, multi-task, and efficiently adaptable—that consolidates the network's cognitive capabilities into a unified backbone from which specialized, deployment-ready variants can be derived. The second is a multi-agent system that operationalizes this intelligence as coordinated autonomous action across both the radio access and core network domains, supported by a digital twin for experiential learning and a knowledge graph for structured reasoning. These pillars pave the way from today's human-as-doer network operations toward a 6G paradigm in which humans serve as supervisors of a largely self-managing infrastructure. %Realizing this vision, however, demands commensurate advances in security and privacy—particularly in defending the foundation model against cross-task poisoning and knowledge extraction, protecting location privacy under pervasive ISAC sensing, and establishing dynamic trust and access control frameworks suited to autonomous agent coordination. The challenges are substantial, but so is the opportunity: a network that is simultaneously more resilient, more autonomous, and more secure represents not merely a generational upgrade in connectivity, but a qualitative expansion in what communication infrastructure can do for society.

\newpage
\balance
\bibliographystyle{ACM-Reference-Format}
\bibliography{sample-base}

\end{document}